\documentclass[10pt,twocolumn,letterpaper]{article}

\usepackage[pagenumbers]{iccv} 
\definecolor{iccvblue}{rgb}{0.21,0.49,0.74}
\usepackage[pagebackref,breaklinks,colorlinks,allcolors=iccvblue]{hyperref}
\usepackage{url}            
\usepackage{booktabs}       
\usepackage[margin=1in]{geometry}
\usepackage{amsfonts}       
\usepackage{nicefrac}       
\usepackage{microtype}      
\usepackage{xcolor}         
\usepackage{float}
\usepackage{scalerel}
\usepackage{tabularx}
\usepackage{multirow}
\usepackage{adjustbox}
\usepackage{amsmath}
\usepackage{stfloats}
\usepackage{colortbl}
\usepackage{rotating}

\def\paperID{10406} 
\def\confName{ICCV}
\def\confYear{2025}

\title{ProtoGS: Efficient and High-Quality Rendering with 3D Gaussian Prototypes}

\author{Zhengqing Gao$^{1}$
\and
Dongting Hu$^{2}$
\and
 Jia-Wang Bian$^{1}$
\and
 Huan Fu$^{3}$
\and
 Yan Li$^{1}$
\and
 Tongliang Liu$^{1,4}$
\and
Mingming Gong$^{1,2,*}$
\and
Kun Zhang$^{1,5,*}$
\and
{\small $^{1}$Mohamed bin Zayed University of Artificial Intelligence}
{\small $^{2}$University of Melbourne}
{\small $^{3}$Alibaba Group}
\\
{\small $^{4}$The University of Sydney}
{\small $^{5}$Carnegie Mellon University}
}

\begin{document}
\twocolumn[{
\renewcommand\twocolumn[1][]{#1}
\maketitle
\begin{center}
    \captionsetup{type=figure}
    \includegraphics[width=0.99\linewidth]{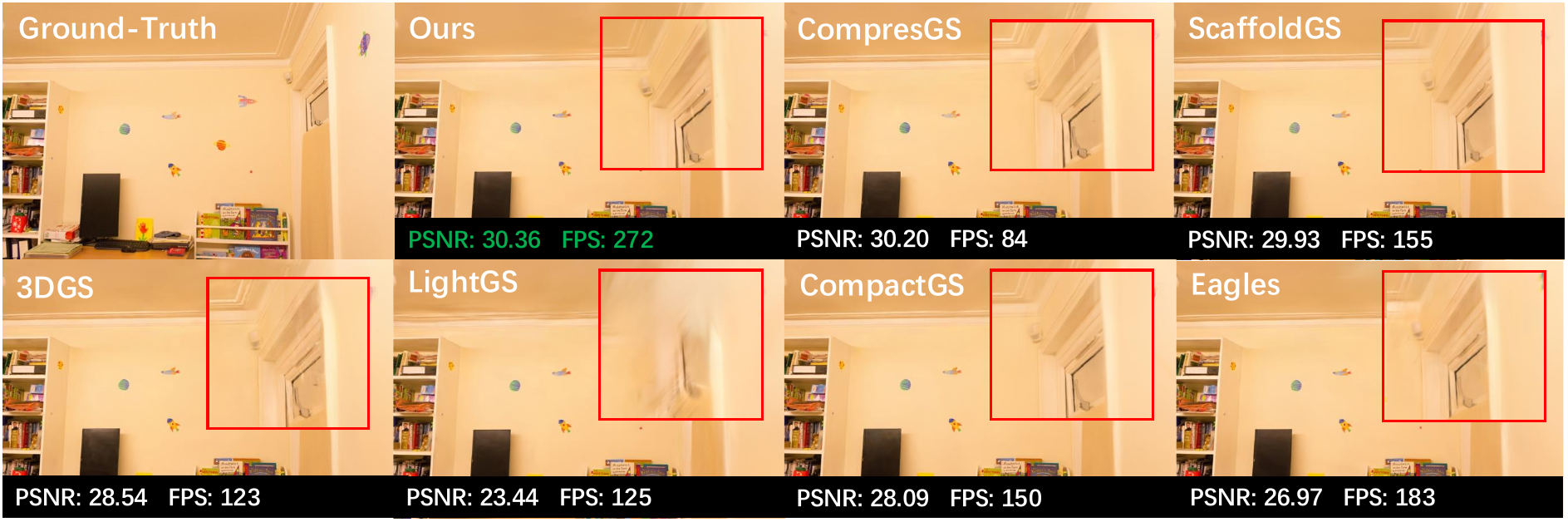}
    \captionof{figure}{We propose a novel method that significantly reduces the memory footprint of 3D Gaussian Splatting, meanwhile enjoys excellent visual quality and high rendering speed. Please zoom in for better clarity, and more results are in supplementary material.}
\end{center}
}]
\begin{abstract}

3D Gaussian Splatting (3DGS) has made significant strides in novel view synthesis but is limited by the substantial number of Gaussian primitives required, posing challenges for deployment on lightweight devices. Recent methods address this issue by compressing the storage size of densified Gaussians, yet fail to preserve rendering quality and efficiency. To overcome these limitations, we propose ProtoGS to learn Gaussian prototypes to represent Gaussian primitives, significantly reducing the total Gaussian amount without sacrificing visual quality. Our method directly uses Gaussian prototypes to enable efficient rendering and leverage the resulting reconstruction loss to guide prototype learning. To further optimize memory efficiency during training, we incorporate structure-from-motion (SfM) points as anchor points to group Gaussian primitives. Gaussian prototypes are derived within each group by clustering of K-means, and both the anchor points and the prototypes are optimized jointly. Our experiments on real-world and synthetic datasets prove that we outperform existing methods, achieving a substantial reduction in the number of Gaussians, and enabling high rendering speed while maintaining or even enhancing rendering fidelity.
\end{abstract}

\footnote{$^{*}$Equal advising.}

\section{Introduction}

Novel view synthesis is a challenging task in computer vision and graphics, requiring both high rendering quality and efficient processing. Traditional methods \cite{mesh_representation, mesh_review, point_based_rendering} use points or meshes to represent scenes, enabling fast rendering through rasterization techniques. However, these approaches often suffer from limited visual quality, producing artifacts like discontinuities and blurriness that detract from the realism of the rendered scenes.
The revolution of Neural Radiance Fields (NeRFs) \cite{nerf} marked a significant advancement, offering a remarkable rendering quality through neural volumetric rendering, achieving superior visual fidelity. Nevertheless, the vanilla NeRFs' rendering speeds are prohibitively slow, restricting their applications in real-time large-scale scenarios.

Recently, 3D Gaussian Splatting (3DGS) has emerged as an alternative representation in novel view synthesis, praised for its rapid rendering speed and high-quality outputs \cite{3DGS, 3DGS_survey}. However, 3DGS often requires an extensive number of Gaussian primitives to synthesize high-fidelity renderings, resulting in a substantial memory footprint. For example, experiments in the MipNeRF-360 dataset \cite{mipnerf360} indicate that the 3DGS representation of a single scene can occupy approximately 800MB of hardware memory. To reduce such 3DGS memory demands, several compression techniques \cite{CompactGS, LightGS, CompressedGS, ScaffoldGS, eagles} have been proposed. One strategy involves cutting or distilling 3DGS in a post hoc manner \cite{LightGS, CompressedGS}. Although this reduces memory requirements, maintaining high rendering quality still requires a substantial number of Gaussian primitives, resulting in only limited improvements in rendering efficiency.

Another strategy involves hybrid representations \cite{eagles, ScaffoldGS, CompactGS}, combining MLP-based rendering with 3DGS. Although this approach improves memory efficiency, it introduces additional computational overhead from MLP decoding, resulting in a reduced rendering speed. Together, these limitations highlight the ongoing challenge of balancing rendering quality, memory efficiency, and computational cost in 3DGS rendering.

To address these limitations, we designed a novel method named Proto-GS that substantially reduces the number of primitives while preserving high rendering quality and efficiency. Our method is inspired by the observation that Gaussian primitives in nearby regions, particularly homogeneous ones, often exhibit similarity. We propose learning Gaussian prototypes to represent multiple Gaussian primitives in nearby regions. Each prototype captures the shared features of surrounding primitives, thereby reducing the number of primitives. 
\label{sec:two_stage}

Deriving prototypes from primitives requires determining which primitives should be assigned to a given prototype, which is computationally intensive. An intuitive but naive solution is to learn prototypes in a two-stage manner, where primitives are merged into prototypes derived by clustering K-means in the first stage, and then prototypes are optimized to improve rendering quality at the second stage. However, such a strategy is prone to lose texture and geometry information because it does not take into account rendering loss when deriving prototypes. Moreover, optimizing prototypes whose attributes are damaged might lead to local optimal solutions and, as a result, to poor visual quality. 

To address above issues, we employ rendering loss to provide guidance to the K-means clustering process, and learn a more efficient representation with high fidelity. To further reduce the memory footprint, we incorporate SfM points as anchors, maintaining an anchor bank to group Gaussian primitives into multiple tiles according to their location. The prototypes are then derived within each tile sequentially. Using the joint optimization process and the SfM anchoring mechanism, we efficiently reduce the redundancy of primitives, meanwhile, preserving high visual quality. Compared with previous works, ProtoGS avoids representing scenes with implicit MLPs, so that enjoys higher rendering efficiency. Extensive experiments are performed on various indoor, outdoor, and synthetic datasets to demonstrate the superiority of our method in memory and rendering efficiency.

In summary, our contributions are as follows.
\begin{itemize}
    \item We condense redundant homogeneous primitives of 3DGS into compact prototypes that efficiently preserve important visual characteristics.
    \item We propose an efficient strategy to derive Gaussian prototypes by tile-wise rendering-guided  K-means iteratively without destroying the original geometry and texture information.
    \item We leverage the anchor bank together with conventional rendering loss to provide guidance to encourage robust derivation of Gaussian prototypes.
\end{itemize}
\section{Related Work}

\paragraph{Novel View Synthesis}
NeRF \cite{nerf} uses MLPs to learn volumetric features from multi-view images and applies volumetric rendering to produce view-dependent outputs. In addition, other methods \cite{scaffoldgs_33, scaffoldgs_34, scaffoldgs_46, scaffoldgs_54} also employ MLPs to predict features of surface points. Thanks to the powerful representation capability of MLPs and the volumetric rendering pipeline, these methods achieve impressive rendering quality in novel view synthesis. However, despite their excellent performance, these approaches face challenges such as slow rendering speeds due to extensive point sampling along numerous camera rays. To overcome this limitation, several approaches \cite{compactgs_7, compactgs_8, compactgs_11, compactgs_34, compactgs_35} have been proposed to accelerate training and rendering speeds, although with a slight trade-off in visual quality. Additionally, some methods \cite{scaffoldgs_15, instant_ngp, grid_based2, scaffoldgs_10, scaffoldgs_12, compactgs_11} use dense voxel grids within multiple spatial grids to represent scenes. In particular, Instant-NGP \cite{instant_ngp} introduced a hash grid structure, which enables much faster feature retrieval, achieving real-time rendering of neural radiance fields. Plenoxel \cite{grid_based2} improves training and rendering speeds by using a sparse voxel grid to interpolate within a continuous density field. Furthermore, \cite{scaffoldgs_10} applies tensor factorization to improve storage and rendering efficiency, while \cite{compactgs_11} uses neural representations to reconstruct a 3D scene with the flexibility to add a temporal plane to capture dynamic elements. Despite these advancements, efficient and high-quality rendering remains an ongoing challenge.

Point-based methods \cite{scaffoldgs_53, scaffoldgs_48, 3DGS} reconstruct 3D scenes using geometric primitives and leverage rasterization techniques during rendering, thus achieving high rendering speeds. Specifically, \cite{scaffoldgs_48} employs volume rendering along with region growth and point pruning to reconstruct fine-grained scenes, while \cite{scaffoldgs_53} introduces differentiable surface splatting to prevent artifacts during the rendering process. However, these methods still suffer from sophisticated ray-marching operations.

\paragraph{Compression for 3DGS} 3DGS begins with SfM points and creates additional finely segmented ellipsoid primitives during differentiable optimization to explicitly represent 3D scenes. Although it achieves high visual quality and rendering speed, 3DGS requires substantial storage because it tends to generate a large number of redundant primitives. To address this issue, various methods have been proposed. \cite{CompactGS, LightGS, CompressedGS, eagles, reducedGS} employ quantization \cite{VQ-VAE} or distillation to represent features in low-dimensional spaces and reduce the number of primitives using learnable masks or importance scores. Other approaches \cite{ScaffoldGS, octreegs, CompGS, ContextGS} introduce neural Gaussians derived from a few anchor points and the corresponding offset parameters to reconstruct 3D scenes with lower storage costs; these anchor points can be densified and pruned during optimization to maintain visual quality.
Furthermore, Gaussian primitives are embedded in 2D grids introduced by \cite{self-organize}, which also utilizes JPEG XL to further reduce the storage size of low-dimensional data. Although these methods reduce the memory footprint of 3DGS, the rendering efficiency is undermined by the inference process of neural representations and the decompression required to recover condensed information.
\section{Method}

In this section, we start with an overview of 3DGS, then introduce how to employ SfM anchoring to reduce memory footprint. We specifically emphasize how we derive 3D Gaussian prototypes using the iterative K-means algorithm guided by a conventional rendering loss.

\subsection{Preliminary: 3D Gaussian Splatting}
\label{sec:3DGS}

As an explicit representation method, 3DGS \cite{3DGS} begins with SfM points extracted from multi-view images of the scene \cite{SfM_points}. 3DGS creates anisotropic Gaussian primitives, whose geometry is defined by
\begin{equation}
\label{def:gaussian_primitives}
    G(x)=e^{-\frac{1}{2}(x-h)^{T}\sum^{-1}(x-h)},
\end{equation}
where $\Sigma$ is the full 3D world space covariance matrix, $h$ denotes the center of primitives, and each primitive is decomposed into learnable parameters rotation and quaternion centered by location $x$, texture and color are modeled by scaling coefficients, opacity, and spherical harmonics (SHs). During rendering, view-dependent images are reconstructed from the Gaussian primitives using rasterization \cite{rasterization} and splatting \cite{splatting} techniques. The 
 pixel value of a rendered image $\mathcal{I}_{j,k}$ is obtained by
 \begin{equation}
 \label{render_eq}
     \mathcal{I}_{j,k}=\sum_{i=1}^{N}\alpha_{i}c_{i}G^{(j,k)}(x),
 \end{equation}
 where $\alpha_i$ and $c_{i}$ are the color and opacity. Together with rotation quaternion and scaling coefficients, we denote the dimension of these data as $d$. Suppose the number of primitives is $N$, we can represent all primitives as $G\in R^{N\times d}$. The reconstruction error provides supervision that guides the parameter updates for $G$.

Given the sparse nature of SfM points, a densification stage is crucial to increase the number of primitives, following specific criteria. In over-reconstructed areas, primitives are split to enhance detail, while in under-reconstructed areas, primitives are cloned to add one additional primitive. This adaptive densification helps achieve a more comprehensive scene representation.

\subsection{SfM Anchoring}
\label{sec:SfM_anchoring}
Deriving prototypes from all primitives requires huge memory because we have to determine which prototype to assign given a primitive. One intuitive measure to alleviate the above issue is splitting all primitives into multiple tiles according to their location, and then derive prototypes within each tile one by one. However, such a splitting operation might put primitives that are consistent in geometry and texture into different tiles, consequently generating damaged prototypes and decreasing the rendering quality. In addition, it also fails to take into account the signal from the image-reconstruction error. 

To enable the derivation on limited computation resources while reserving texture and geometry features of scenes, we propose the SfM anchoring strategy to divide primitives into multiple tiles before deriving prototypes. 

\begin{figure*}
  \centering
  \includegraphics[width=16.5cm]{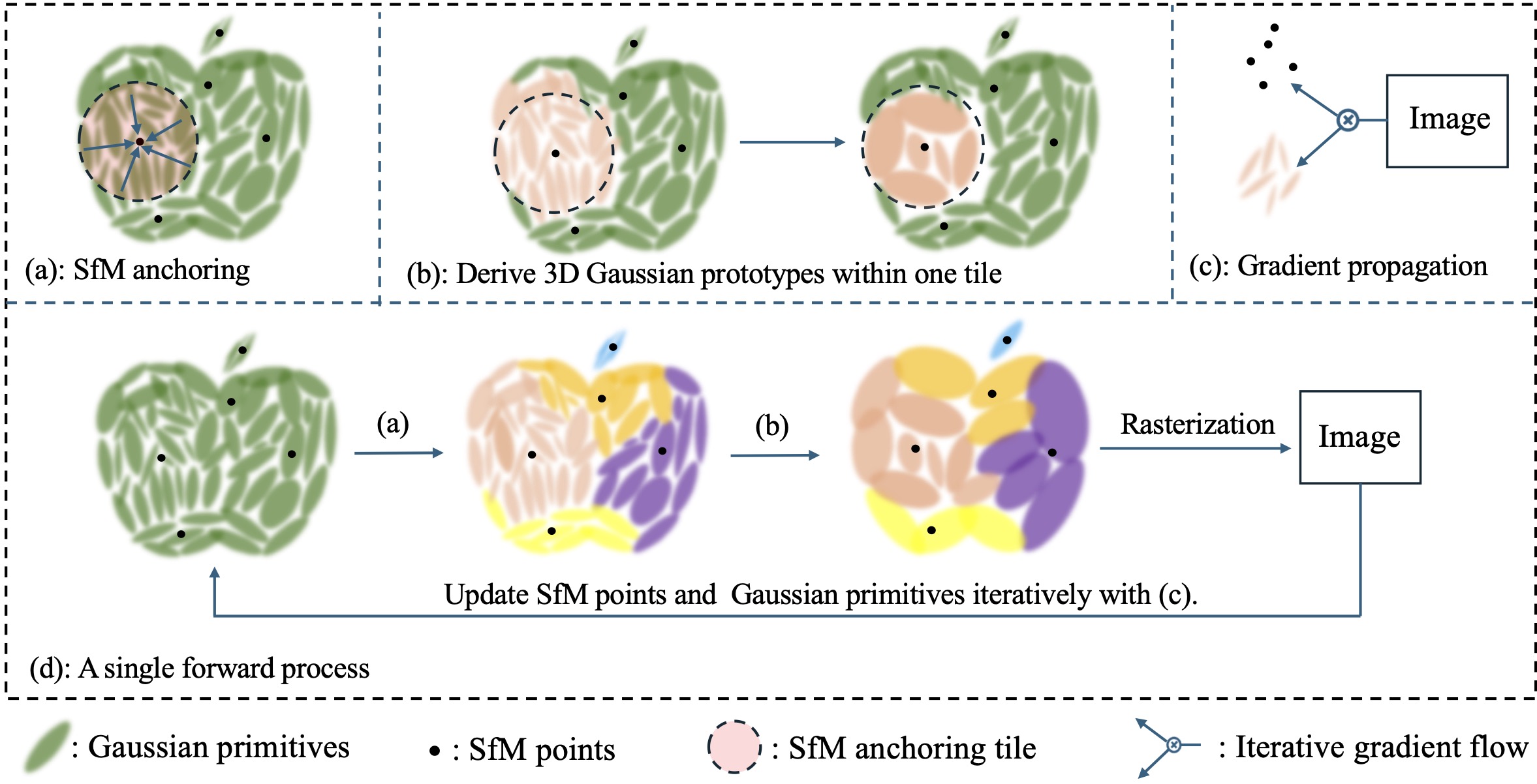}
  \caption{Overview of our method. Starting with Gaussian primitives (denoted by green ellipsoids), we first assign primitives to multiple tiles (distinguished by different colors) centered by SfM points as shown in \textbf{(a)}, then we derive 3D Gaussian prototypes from primitives within each tile in \textbf{(b)} controlled by a compression ratio. After that, we utilize rasterization to render images based on prototypes, and propagate gradients back to update Gaussian primitives iteratively as shown in \textbf{(d)}. After multiple processes \textbf{(d)}, we replace current primitives with prototypes to reduce the total number of primitives.}
  \label{overview}
\end{figure*}

We denote the location of SfM points by coordinates in the spatial coordinate system as $\mathcal{P}\in \mathbb{R}^{q\times 3}$, where $q$ is the number of SfM points. Because the number of SfM points is large, we randomly select $M$ points from $\mathcal{P}$, and denote as $P\in\mathbb{R}^{M\times 3}$. Then we assign each $G_{i}\in G$ to a tile centered by one anchor point $P_{m}\in P$ if 
\begin{equation}
m= \arg\min_{t} D(G_{i}, P_{t}),
\end{equation}
where $D(\cdot,\cdot)$ represents the Euclidean distance between an anchor point $P_t$ and the first three dimensions of $G_i$. In other words, we simply assign $G_{i}$ to the tile centered by the 
closest SfM point in the spatial coordinate system as shown in Fig. \ref{overview} (a). As a result, we ensure that prototypes are derived on the basis of primitives belonging to the same tile. Moreover, this operation prevents one primitive from being merged into prototypes that are far from it. In addition, the initial SfM points are computed by COLMAP, which might generate noisy results, while small portion of randomly selected SfM points can rule out some bad points. Based on SfM anchoring tiles we derive prototypes on those tiles sequentially with computational efficiency gurantee.

\subsection{Derivation of 3D Gaussian Prototypes}
 Given primitives $G\in R^{N\times d}$, which are divided into $M$ tiles as introduced in \ref{sec:SfM_anchoring}, prototypes are derived within each tile sequentially as shown in Fig. \ref{overview} (b). We denote primitives belonging to the $m$-th tile as $G_{m}$ containing $n_{m}$ primitives, where $m\in\{1, 2, ..., M\}$ and $\sum_{m=1}^{M}n_{m}=N$. For each primitive $G_{m,i}$ in the $m$-th tile, we introduce a corresponding set of binary indicator variables $r_{i,k}^{m}$, where $k\in\{1,2,...,K^{m}\}$ describing which of the $K^{m}$ clusters the primitive is assigned to, so that if $G_{m,i}$ is assigned to cluster $k$, then $r_{i,k}^{m}=1$, and $r_{i,j}^{m}=0$ for $j\neq k$. We hope to derive $K^{m}$ prototypes in the $m$-th tile, so we define an objective function given by
\begin{equation}
    J^{m}=\sum_{i=1}^{n_{m}}\sum_{k=1}^{K^{m}}r_{ik}^{m}\Vert G_{m,i}-\boldsymbol{\mu}_{m,k}\Vert^{2},
\end{equation}
which represents the sum of the squares of the distances of each primitive to its assigned vector $\mu_{m,k}$, serving as prototypes derived from the $m$-th tile. Our goal is to find values for the $\{r_{ik}^{m}\}$ and the $\{\mu_{m,k}\}$ so as to minimize $J^{m}$. Then we write the objective function on all tiles as 
\begin{equation}
    \mathcal{L}_{c}=\sum_{m=1}^{M}J^{m}.
\end{equation}
We simply assign the primitive $G_{m,i}$ to the nearest cluster center. More formally, this can be expressed as 
\[
r_{ik}^{m} = 
\begin{cases} 
1 & \text{if } k = \arg\min_j \|G^{m,i} - \boldsymbol{\mu}_{m,j}\|^2, \\
0 & \text{otherwise.}
\end{cases}
\]

Well-derived prototypes should not only merge as many homogeneous primitives within the same tile as possible but also guarantee high visual quality in the rendering process. We design the objective function as
\begin{equation}
    \label{loss_func}
    \mathcal{L}=(1-\lambda)\mathcal{L}_{1}(I,\hat{I})+\lambda\mathcal{L}_{\text{D-SSIM}}(I, \hat{I})+\lambda_{\text{c}}\mathcal{L}_{\text{c}},
\end{equation}
where $\mathcal{L}_{1}(I,\hat{I})$ and $\mathcal{L}_{D-SSIM}(I,\hat{I})$ are defined as
\begin{equation}
    \begin{cases}
        &\mathcal{L}_{1}(I,\hat{I})=| I-\hat{I}|,\\
    &\mathcal{L}_{D-SSIM}(I,\hat{I})=1-SSIM(I, \hat{I}),
    \end{cases}
\end{equation}
and both terms take ground-truth image ${I}$ and rendered images $\hat{I}$ as input and evaluate the visual difference between those images. The rendered image is obtained from \textbf{prototypes} \eqref{render_eq}, i.e.,
$\hat{I} = f(\boldsymbol{\mu})$, where $\mu$ denotes all the prototypes and $f(\cdot)$ denotes the rendering function.
This objective provides supervision to update the primitives, SfM anchors, and also the primitive to prototype assignment $r_{ik}^{m}$, as shown in Fig. \ref{overview} (c). At the end, we reduce primitives by replacing them with prototypes after multiple updating processes (d) in Fig. \ref{overview}. 
\section{Experiments}

\begin{table*}[!htbp]
\centering
\caption{Quantitative evaluation of our method compared to other works based on \textbf{PSNR}$\uparrow$, \textbf{SSIM}$\uparrow$, \textbf{LPIPS}$\downarrow$, and \textbf{average }\textbf{training time}$\downarrow$ and \textbf{FPS}$\uparrow$, over five real-world and synthetic datasets. We highlight the best three values in different colors for each metric.}
\label{tab:comparison1}
\renewcommand{\arraystretch}{1.2} 
\resizebox{\textwidth}{!}{
\begin{tabular}{lccccccccccccccccc} 
\toprule
 \textbf{Dataset}& \multicolumn{3}{c}{\textbf{Deep Blending}} & \multicolumn{3}{c}{\textbf{DTU}} & \multicolumn{3}{c}{\textbf{Mip-NeRF 360}} & \multicolumn{3}{c}{\textbf{Tanks\&Temples}} & \multicolumn{3}{c}{\textbf{NeRF-Synthetic}} & \multicolumn{2}{c}{\textbf{Avg. Speed}}\\
 \cmidrule(lr){2-4} \cmidrule(lr){5-7} \cmidrule(lr){8-10} \cmidrule(lr){11-13} \cmidrule(lr){14-16} \cmidrule(lr){17-18}
 \textbf{Methods}& \textbf{PSNR}  & \textbf{SSIM}  & \textbf{LPIPS}  & \textbf{PSNR}  & \textbf{SSIM}  & \textbf{LPIPS}  & \textbf{PSNR}  & \textbf{SSIM}  & \textbf{LPIPS}  & \textbf{PSNR}  & \textbf{SSIM}  & \textbf{LPIPS}  & \textbf{PSNR}  & \textbf{SSIM}  & \textbf{LPIPS}  & \textbf{Train} & \textbf{FPS}\\
\midrule
Eagles \cite{eagles} & 28.58 & \cellcolor{orange!45}0.904 & \cellcolor{pink}0.244 & \cellcolor{yellow!45}28.54 & \cellcolor{orange!45}0.889 & \cellcolor{pink}0.286 & 27.23 & 0.810 & 0.240 & 23.02 & 0.830 & 0.200 & 32.53 & 0.964 & 0.040 & 25m &\cellcolor{yellow!45}186\\
CompresGS \cite{CompressedGS} & 29.46 & 0.899 & \cellcolor{yellow!45}0.258 &28.36 & \cellcolor{yellow!45}0.883 & 0.297& 27.11 & 0.801 & 0.243 &21.87 & 0.799 & 0.248 & 31.09 & 0.958 & 0.052 &30m &85\\
CompactGS \cite{CompactGS} & \cellcolor{orange!45}29.79 & 0.900 & 0.258 & 28.06 & 0.879 & 0.296 & 27.08 & 0.798 & 0.247 & \cellcolor{yellow!45}23.32 & \cellcolor{yellow!45}0.831 & \cellcolor{yellow!45}0.201 & \cellcolor{yellow!45}33.33 & 0.936 & 0.062 & 25m &\cellcolor{orange!45}199\\
ScaffoldGS \cite{ScaffoldGS} & \cellcolor{pink}30.21 & \cellcolor{pink}0.906 & \cellcolor{yellow!45}0.254 & \cellcolor{orange!45}28.99 & \cellcolor{yellow!45}0.883 & 0.307 & \cellcolor{orange!45}28.50 & \cellcolor{orange!45}0.842 & \cellcolor{orange!45}0.220 & \cellcolor{pink}23.96 & \cellcolor{pink}0.853 & \cellcolor{pink}0.177 & \cellcolor{pink}33.68 & 0.928 & 0.071 & \cellcolor{yellow!45}22m &165\\
LightGS \cite{LightGS}& 29.30 & 0.895 & 0.273 & 27.42 & 0.873 & 0.312 & 27.16 & 0.803 & 0.246 & 23.09 & 0.820 & 0.226 & 32.18 & \cellcolor{yellow!45}0.965 & \cellcolor{yellow!45}0.037 & \cellcolor{orange!45}21m &\cellcolor{yellow!45}186\\
3DGS \cite{3DGS}& 29.58 & \cellcolor{orange!45}0.904 & \cellcolor{pink}0.244 & 28.45 & \cellcolor{yellow!45}0.883 & \cellcolor{yellow!45}0.291 & \cellcolor{yellow!45}27.50 & \cellcolor{yellow!45}0.813 & \cellcolor{yellow!45}0.221 & 21.99 & 0.807 & 0.236 & 33.21 & \cellcolor{pink}0.969 & \cellcolor{pink}0.031 & \cellcolor{pink}20m &125\\
\midrule
ProtoGS (Ours) & \cellcolor{yellow!45}29.71 & \cellcolor{yellow!45}0.902 & \cellcolor{orange!45}0.248 & \cellcolor{pink}31.19 & \cellcolor{pink}0.891 & \cellcolor{orange!45}0.290 & \cellcolor{pink}28.73 & \cellcolor{pink}0.861 & \cellcolor{pink}0.199 & \cellcolor{orange!45}23.59 & \cellcolor{orange!45}0.843 & \cellcolor{orange!45}0.184 & \cellcolor{orange!45}33.46 & \cellcolor{orange!45}0.968 & \cellcolor{orange!45}0.033 & 32m &\cellcolor{pink}225\\
\bottomrule
\end{tabular}
}
\end{table*}

We begin by presenting the datasets and numerical metrics used to evaluate our method. Next, we detail the implementation aspects of our experiments and compare the numerical results of our method with those of other competitors across various datasets. Finally, we emphasize the importance of each component of our approach through an ablation study.

\subsection{Dataset and Metrics}

We conducted experiments on diverse open-source datasets to evaluate our method's performance. Specifically, we tested on the same scenes used in previous studies such as 3DGS \cite{3DGS} and ScaffoldGS \cite{ScaffoldGS}, including seven scenes from the Mip-NeRF360 dataset \cite{mipnerf360}, two scenes from Tanks \& Temples \cite{tanks_temple}, two scenes from DeepBlending \cite{deep_blending} and the synthetic Blender dataset \cite{nerf}. In addition, we evaluated all methods on 15 scenes from the DTU dataset \cite{DTU}, which includes varied illumination conditions.
For quality assessment of the rendered images, we use PSNR \cite{PSNR}, SSIM \cite{SSIM}, and LPIPS \cite{LPIPS}. To evaluate rendering and compression efficiency, we measure frames per second (FPS) and storage size (MB). In addition, we report the average running time across all datasets, with detailed time for each dataset available in the supplemental materials.

\subsection{Baseline and Implementation}

We selected various 3D scene representation methods as baselines to compare with our method. Specifically, we evaluated the following methods: Eagles \cite{eagles}, which leverages quantization to represent features in discrete latent space and prune redundant quantized primitives; CompactGS \cite{CompactGS}, which reduces Gaussian points by learning masks for each primitive and compresses color features with grid-based neural fields; ScaffoldGS \cite{ScaffoldGS}, which learns a compressed hybrid representation based on Gaussian anchor points and feature augmentation with MLPs; LightGS \cite{LightGS}, which prunes insignificant Gaussian points to reduce primitives and distills spherical harmonics (SHs) into a lower-degree compact representation; and 3DGS \cite{3DGS}, an explicit representation method that utilizes the redundant primitives alongside rasterization techniques to enable efficient rendering.

All experiments, including re-running competitor methods to reproduce their results, were conducted on a machine with one NVIDIA A100 (40GB) GPUs. For our method, we set $\lambda_{c}=0.0001$ across all experiments, while other parameters in the loss function (Equation \ref{loss_func}) remain consistent with those used in 3DGS \cite{3DGS}. When iteratively deriving prototypes (shown in part (c) of Figure \ref{overview}), we set the interval $t=100$, meaning Gaussian primitives are optimized over 100 iterations before fine-tuning all SfM points in a single iteration. Additionally, we apply a progressive decay rate of 0.5, which halves the number of 3D Gaussian prototypes after some iterations.

\subsection{Results Analysis}
We analyze the rendering quality and efficacy of our method from qualitative and quantitative perspectives. We also demonstrate our effectiveness in reducing redundant finely-chopped primitives by ellipsoids visualization. Additionally, we report numerical results of compression and rendering efficiency in storage size and FPS.

\paragraph{Quantitative Comparison} We present quantitative results in Table \ref{tab:comparison1}. The quality metrics of the baseline methods are consistent with those reported in their own papers. We show that our approach achieves comparable or higher rendering quality as ScaffoldGS and 3DGS in Deep Blending \cite{deep_blending}, DTU \cite{DTU}, Mip-NeRF360 \cite{mipnerf360}, and Tanks\&Temples \cite{tanks_temple}. Our method enjoys the most competitive rendering speed due to the superiority of our compression strategy. For more detailed numerical results across other scenes, see the supplemental materials. In addition, we specifically compare our method with CompactGS \cite{CompactGS} and LightGS \cite{LightGS}, we render images with primitives selected by them. As shown in Table \ref{tab:num_prototypes_comparison}, the results evaluated on the NeRF-Synthetic dataset show that our method achieves better rendering performance with much fewer primitives.\\

\begin{table}[htbp]
\centering
\caption{Quantitative comparison of our method with other competitors based on \textbf{PSNR}$\uparrow$ and \textbf{the number of primitives}$\downarrow$ on the NeRF-Synthetic dataset.}
\label{tab:num_prototypes_comparison}
\begin{tabularx}{0.4\textwidth}{lcr}
\toprule
\textbf{Dataset} & \multicolumn{2}{c}{NeRF-Synthetic}\\

\textbf{Method} &\textbf{ PSNR } & \textbf{\#Primitives}   \\
\midrule

CompactGS \cite{CompactGS} & 33.33 & 160K \\
3DGS \cite{3DGS} & 33.21 & 261K \\
LightGS\cite{LightGS} & 32.18 & 90K \\
\midrule
ProtoGS (Ours) & 33.45 & 45K\\
\bottomrule
\end{tabularx}
\end{table}

\begin{figure*}
    \centering
    \includegraphics[width=16cm, height=20cm, keepaspectratio=false]{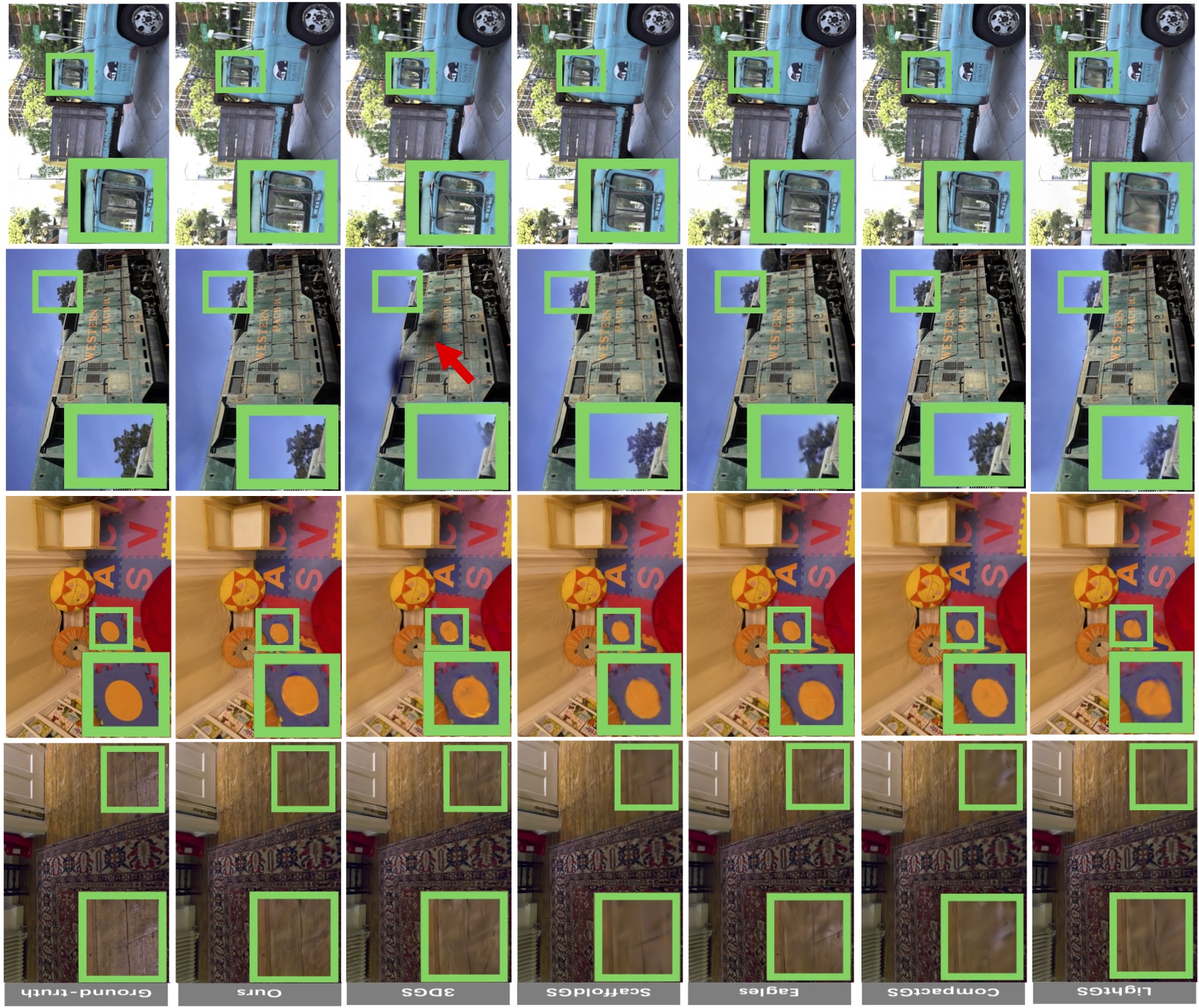}
    \caption{From the second to the last row, we show qualitative comparison of ProtoGS (Ours), 3DGS, ScaffoldGS, Eagles, CompactGS, and LightGS . We show the corresponding ground truth images on the first row, from the left to 
right: Drjohnson, Playroom from the deep blending dataset; and Truck and Train from Tanks\&Temples. Visual differences are highlighted by \textcolor{green}{green frames} and \textcolor{red}{red arrow} and corresponding regions are zoomed in and put in the bottom left corner of each image.}
    \vspace{2cm}\label{fig:qualitative_comparison}
\end{figure*}

\paragraph{Qualitative Comparison}
We provide a qualitative comparison of our method with 3DGS\cite{3DGS}, ScaffoldGS \cite{ScaffoldGS}, Eagles \cite{eagles}, CompactGS \cite{CompactGS}, and LightGS \cite{LightGS} in various scenes in Figure \ref{fig:qualitative_comparison}. This shows that our method achieves visual quality comparable to or higher than that of other competitors. We showcase images rendered on (from top to bottom rows) (1) DB-Drjohnson, (2) DB-playroom, (3) T\&T-Train, and (4) T\&T-Truck. These scenes cover challenging indoor scenes and outdoor scenes. We use \textcolor{green}{green frames} to highlight regions where our method achieves higher visual quality than others. We also use \textcolor{red}{red arrow} to highlight the area that 3DGS cannot reconstruct due to floating. The above visual results show that our method achieves higher visual quality and avoids generating floaters.

\begin{figure}
    \centering
    \includegraphics[width=0.475\textwidth]{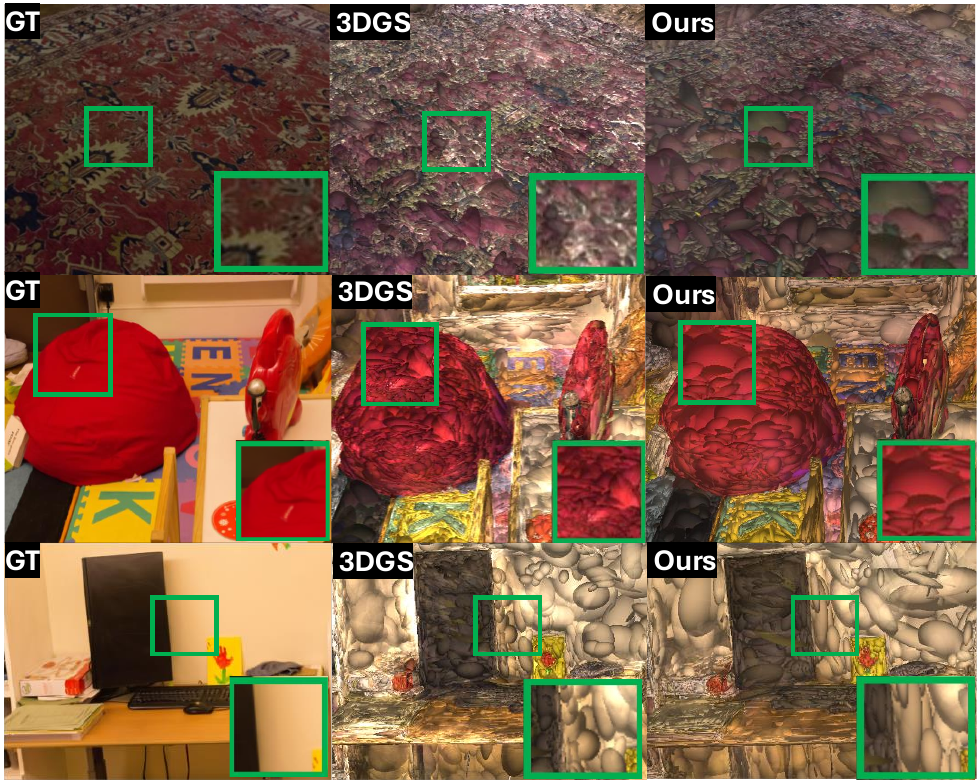}
    \caption{Visualization of ellipsoids. We visualize ellipsoids learned by our method and 3DGS on Drjohnson (first row) and Playroom scenes (second and third row). We use \textcolor{green}{green frames} to show that our method successfully circumvents redundant finely chopped primitives like 3DGS, and the zoomed-in regions are shown in the bottom right corners of each image.}
    \label{fig:visualized_prototypes}
\end{figure}

\paragraph{Prototypes Visualization} A visualization of 3D Gaussian prototypes and ellipsoids is shown in Figure \ref{fig:visualized_prototypes}. We render the Drjohnson and Playroom scenes by our approach and 3DGS, respectively. It can be observed that our approach uses fewer compact prototypes (middle) but not finely chopped ellipsoids created by 3DGS (right) to reconstruct scenes, as highlighted by \textcolor{green}{green frames} in Figure \ref{fig:visualized_prototypes}. Especially in textureless areas (for example, the red cushion), our approach needs much fewer prototypes to achieve equal rendering quality as 3DGS.

\begin{figure} 
    \centering
        \includegraphics[width=0.45\textwidth]{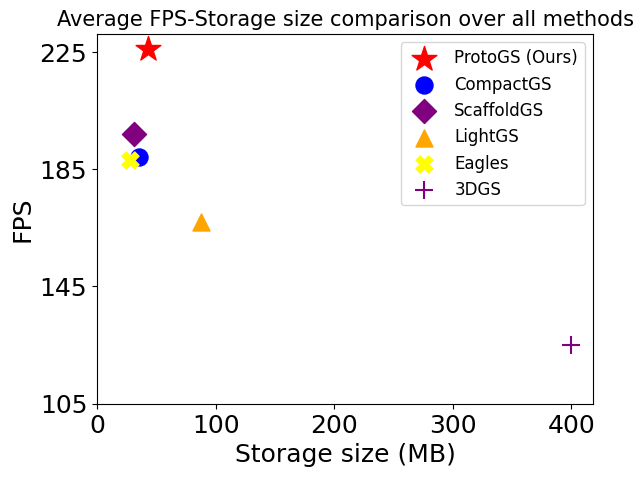}
        \caption{Comparison of our method with other competitors in average storage size and rendering speed over all datasets.}
        \label{fig:fps_storage_size}
\end{figure}

\begin{figure}
    
        \centering
        \includegraphics[width=0.47\textwidth]{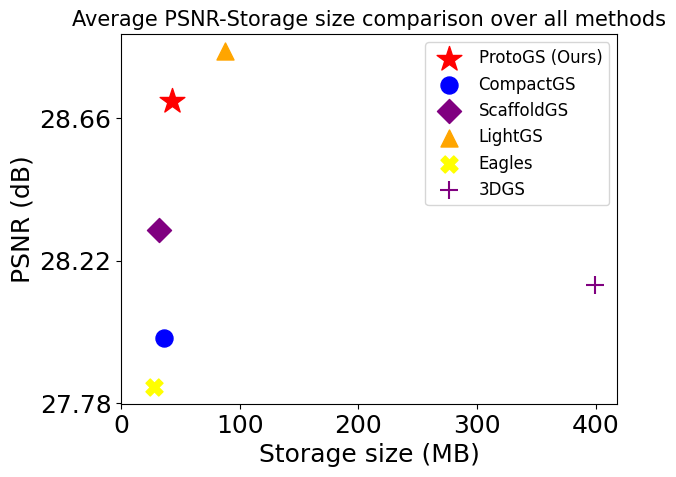}

    \caption{Comparison of our method with other competitors in average visual quality and storage size over all datasets.}
   \label{fig:psnr_storage_size}
\end{figure}

\paragraph{Storage Consumption and Rendering Performance} In Figure \ref{fig:fps_storage_size} and Figure \ref{fig:psnr_storage_size} we give a comprehensive comparison of our method against competing techniques, highlighting our advantages in storage efficiency, rendering speed and visual quality. In Figure \ref{fig:fps_storage_size}, our approach achieves a significant reduction in storage size while achieving the highest rendering speed. In Figure \ref{fig:psnr_storage_size}, our method maintains a compact storage footprint and achieves a high PSNR, demonstrating its ability to deliver high visual quality. These results show that our method offers an optimal balance of reduced storage size and enhanced rendering performance, compared to others. 

\begin{table}[htbp]
\centering
\caption{Quantitative analysis for clustering with rendering guidance on rendering quality, evaluated on the Mip-NeRF360 dataset.}
\label{tab:ablation_study1}
\begin{tabularx}{0.445\textwidth}{lccc}
\toprule
clustering & \multicolumn{3}{c}{Mip-NeRF360}\\

strategy & PSNR  & SSIM  & LPIPS  \\
\midrule

Two stage & ~27.62 & 0.806 & 0.288 \\
Rendering guided & ~28.22 & 0.832 & 0.252 \\

\bottomrule
\end{tabularx}
\end{table}

\subsection{Ablation study}
\begin{figure}
    \centering
    \includegraphics[width=\linewidth]{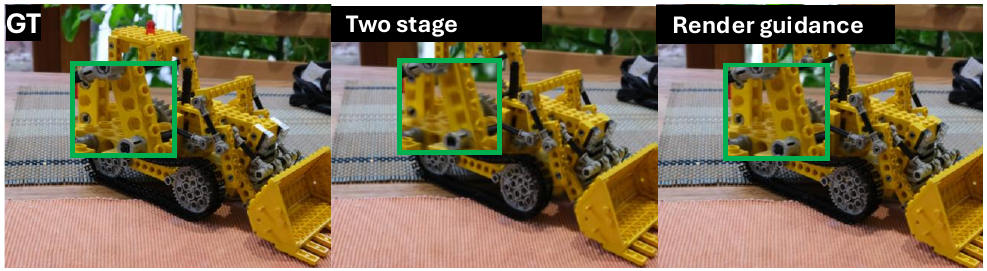}
    \caption{Visual results that show the effectiveness of clustering with rendering guidance.}
    \label{fig:ablation_rendering_guidance}
\end{figure}

\paragraph{Rendering guided clustering} We explored the effectiveness of our rendering-guided clustering on rendering quality, and make comparison with a naive two-stage method mentioned in \ref{sec:two_stage}. We evaluate the visual quality on the Mip-NeRF360 dataset, it can be observed that the visual quality is significantly enhanced by our strategy as shown in Tab. \ref{tab:ablation_study1}, and the visual result is shown in Fig. \ref{fig:ablation_rendering_guidance}.

\begin{figure}
    \centering
    \includegraphics[width=0.85\linewidth]{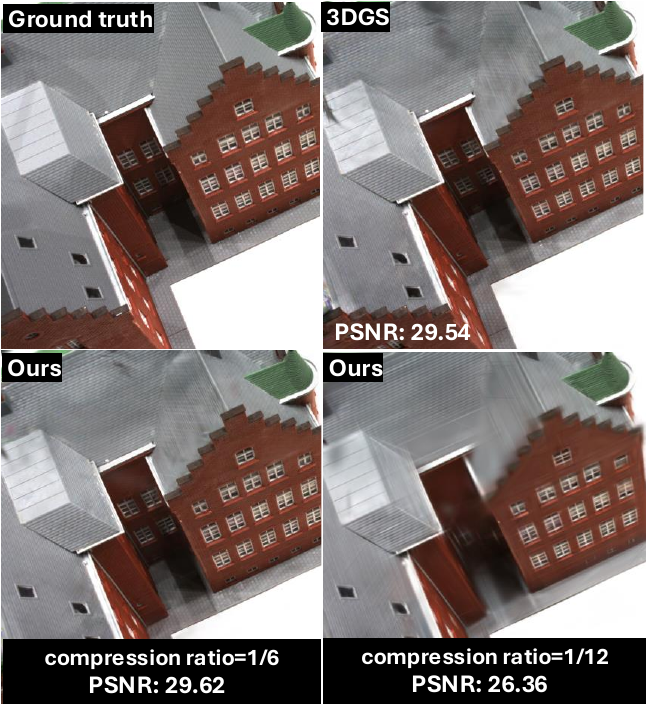}
    \caption{Visual results that show the effects of different compression ratio on visual quality.}
    \label{fig:ablation_compression_ratio}
\end{figure}

\paragraph{Compression ratio} We set the compression ratio to different values to derive prototypes with different amounts. It is obser
ved that an aggressive compression ratio leads to poor visual quality as shown in Fig. \ref{fig:ablation_compression_ratio}.

\paragraph{SfM point amount} We randomly select different amounts of SfM points in the initial stage, according to the Tab. \ref{tab:training_time_comparison}, we can see the amount of initial SfM points does not affect rendering efficiency and quality.
\begin{table}[htbp]

\centering
\caption{Ablation analysis on different numbers of SfM points and time comparison with 3DGS on large-scale playroom scene.}

\begin{tabularx}{0.475\textwidth}{l|c|c|c|c|c}
\hline
\#SfM & 1/5 & 1/2 & 3/4 & 1.00 & 3DGS \\ \hline
PSNR  & 30.08 & 30.02 & 30.08 & 29.98 & 30.93 \\ \hline
Time  & 19.5m & 19.2m & 20.6m & 21.2m & 16.6m \\ \hline
Size  & 57M & 59M & 59M & 60M & 437M \\ \hline
FPS  & 272 & 270 & 270 & 270 & 105 \\ \hline
\end{tabularx}
\label{tab:training_time_comparison}
\end{table}

\subsection{Discussions and Limitations}
Extensive experiments show that our approach enjoys excellent rendering efficiency with 3D Gaussian prototypes derived from primitives, without sacrificing visual quality. Since our prototypes are consistent with Gaussian primitives in terms of representation structure, the efficient rasterization module of 3DGS can be seamlessly applied to our method, enabling us to achieve higher rendering speeds than competitors. However, deriving prototypes is time-consuming, especially when handling large-scale scenes, although this can be mitigated by our controlled progressive decaying technique. We anticipate that our approach can be continuously enhanced to achieve higher performance and help usher in the era of large-scale applications in 3D vision.
\section{Conclusion}
In this work, we propose a promising approach for compact 3D scene representation that enables efficient and high-quality rendering. Our method derives 3D Gaussian prototypes from the Gaussian primitives of 3DGS, constrained by anchor Gaussian primitives based on SfM points. Extensive experiments demonstrate that rendering images with our prototypes achieves higher speeds than other methods without sacrificing visual quality. The superiority of our 3D Gaussian prototypes is evident in their ability to represent repeated, finely segmented Gaussian primitives using fewer, more compact components. Furthermore, we show that Gaussian prototypes can surpass 3DGS in some challenging scenes, underscoring their potential practicality for future AR/VR applications.
{
    \small
    \bibliographystyle{ieeenat_fullname}
    \bibliography{main}
}


\end{document}